\documentclass[12pt]{elsarticle}

\usepackage{amssymb}
\newcommand{\ra}[1]{\renewcommand{\arraystretch}{#1}}
\usepackage{lipsum}
\usepackage{setspace}
\usepackage[ruled,vlined,linesnumbered]{algorithm2e}
\begin{document}

\begin{frontmatter}

%% Title, authors and addresses

%% use the tnoteref command within \title for footnotes;
%% use the tnotetext command for theassociated footnote;
%% use the fnref command within \author or \affiliation for footnotes;
%% use the fntext command for theassociated footnote;
%% use the corref command within \author for corresponding author footnotes;
%% use the cortext command for theassociated footnote;
%% use the ead command for the email address,
%% and the form \ead[url] for the home page:
 \title{Lexicon and Attention based Handwritten Text Recognition System}
 \author[1]{Lalita Kumari\fnref{label2}}
 \ead{lalita@pu.ac.in}
 \address[1]{Department of Computer Science and Applications,Panjab University, Chandigarh, India}
 \author[2]{Sukhdeep Singh\fnref{label3}}
 \ead{sukha13@ymail.com}
 \ead[url]{https://sites.google.com/site/sransingh13/}
 \address[2]{D.M. College(Aff. to Panjab University), Moga, Punjab, India}

\author[3]{VVS Rathore\fnref{label3}}
\ead{vaibhav@prl.res.in}
\address[3]{Computational Services Group, PRL, Ahmedabad, Gujarat, India}

\author[1]{Anuj Sharma\fnref{label4}}
\ead{anujs@pu.ac.in}
\ead[url]{https://anuj-sharma.in}

% \author[1]{Lalita Kumari\corref{cor1}%
% 	\fnref{fn1}}
% \ead{lalita@pu.ac.in}
% \author[2]{Sukhdeep Singh%
% 	\fnref{fn2}}
% \ead{sukha13@ymail.com}
% \ead[url]{https://sites.google.com/site/sransingh13/}
% \author[3]{VVS Rathore%
% 	\fnref{fn3}}
% \ead{vaibhav@prl.res.in}
% 
% \author[1]{Anuj Sharma%
% 	\fnref{fn4}}
% \ead{anujs@pu.ac.in}
% \ead[url]{https://anuj-sharma.in}
% 
%  \affiliation[1]{organization={Department of Computer Science and Applications,
% 		Panjab University},
% 	city={Chandigarh}, 
% 	state={Panjab},
% 	country={India}}
% 
% \affiliation[2]{organization={D.M. College(Aff. to Panjab University, Chandigarh)},
% 	city={Moga}, 
% 	state={Panjab},
% 	country={India}}
% 
%  \affiliation[3]{organization={Computational Services Group, Physical Research Laboratory},
% 	city={Ahmedabad}, 
% 	state={Gujarat},
% 	country={India}}
 
\begin{abstract}
The handwritten text recognition problem is widely studied by the researchers of computer vision community due to its scope of improvement and applicability to daily lives, It is a sub-domain of pattern recognition. Due to advancement of computational power of computers since last few decades neural networks based systems heavily contributed towards providing the state-of-the-art handwritten text recognizers. In the same direction, we have taken two state-of-the art neural networks systems and merged the attention mechanism with it. The attention technique has been widely used in the domain of neural machine translations and automatic speech recognition and now is being implemented in text recognition domain. In this study, we are able to achieve 4.15\% character error rate and 9.72\% word error rate on IAM dataset, 7.07\% character error rate and 16.14\% word error rate on GW dataset after merging the attention and word beam search decoder with existing Flor et al. architecture. To analyse further, we have also used system similar to Shi et al. neural network system with greedy decoder and observed 23.27\% improvement in character error rate from the base model.  
\end{abstract}

\end{frontmatter}

%% \linenumbers

%% main text
\section{Introduction}
One of the major mode of communication is through handwritten text. From the 19\textsuperscript{th} century onwards, there is a rapid growth in the various computer technologies. One such domain is handwriting recognition \cite{dewi2016}. It is sub domain of pattern recognition. The act of transcribing handwritten text into digitized form is named as handwritten text recognition (HTR). This problem is widely studied in the research community due to its omnipresent need where people communicate and interact. The communication mode can be verbal, sign language or handwritten text. At present, handwritten text can be represented in two ways: online and offline. The online handwriting recognition is performed while the text to be recognized is written(e.g. by a pressure setting device), therefore temporal and geometric information is available. It is a way of processing and recognizing the text while writing or entering. The offline handwriting recognition, is performed by collecting handwritten data and fed these data into a computing device for recognition. The HTR task faced various challenges like cursiveness of handwritten text, different size and shape of each character and large vocabularies. There are many 
challenging aspects of any unconstrained handwriting recognition task. For example, in unconstrained handwriting recognition there is no control over author , writing styles and instrument used for writing. Moreover different vertical and horizontal space presents  among different lines or  different words in the same line that causes  uncertainty in total number of  lines and words respectively in a page and line of a handwritten  document for a automatic text recognizer \cite{Kim1999}. 
%###############application of HTR###########################
\par One of the the key application areas of HTR is converting the ancient historical handwritten text into digital form as a part of modern digital library. it helps in bridging the gap among computers and humans in various domains. Apart from it, invoices, notes, accidental claims, feedback forms are usually in the handwritten format and these can be used as digital footprints in industrial domain. Despite from having developed many state of the art techniques , HTR  domain is still far away from having a generic system that is able to read any handwritten text. Initial state of the art  approaches  of HTR domain are based upon Hidden Markov Models (HMMs) \cite{bernard2011,bluche2013,gimenez2014,bunke2010} 
% About CNNS 
Recently, there had been rising in state of the art Neural Network (NN) based techniques, specially Convolutional Neural Network (CNN) techniques for its various challenging tasks. In any typical HTR system an input image  consist of sequence of objects (characters) to be recognized that  contextually depends upon each other. Recognition text length also varies greatly for example, English language word  "To" is of length 2, "Tomorrow" is of length 8 and "The quick brown fox jumps over a lazy dog" is of length 33. The recognition of such sequences of objects is  done either at word or line level. Although word level recognition systems are quite popular but for a generic system it is suboptimal. Firstly, for handwritten text, it is not always feasible to do word segmentation due to their partially overlapped  proximity. In addition to it, for densely written handwritten text, it is complex to detect large number of words and at last there may be scenarios where a word is not separated by a space which is a word separator considered by most of the linguistic systems \cite{Kumari2022}. Thus this study focused on line level recognition architecture to make the system as generic as possible. The  CNN in connection with RNN have been consistently performing good and able to give state of the art results for HTR problem \cite{yenger1990,imagenet2017,shi2017}. The
CNN typically has an convolutional layer that uses convolution operation for extracting the features of the given image by applying various filters. The RNN is used to capture sequences and contextual information hence able to correlate the relationship among characters rather treating them independently. The Bidirectional Long Short-Term Memory (BLSTM) \cite{blstm2015,graves2009} RNN is used to capture long term dependencies and getting more context in comparison  to typical RNN architecture. Deep BLSTM RNN is widely used in speech recognition tasks \cite{graves2014,sak2014}. The output from BLSTM RNN is  character probability vs time matrix, where t\textsubscript{0} is characters probabilities at the start of the sentence and t\textsubscript{n} is characters probabilities of at the end of sentence. In 2006, there was a technique introduced used to train and score  NN architectures in which  input sequence and output labels are given in the form of input and output pair, the Connectionist temporal classification (CTC) function does not depend upon the underlying network architecture \cite{ctc2006}. Hence, in architecture trained by CTC, output of RNN is characters probabilities including blank character. A blank character is a special character introduced in CTC to handle duplicate characters. obtaining the actual text from the output of RNN is called decoding.\par A typical HTR system include various preprocessing steps to minimize the variation of text as much as possible. There are no generic 
preprocessing steps but these usually relies on the input of the HTR system \cite{moya2011}. In this study, we have usued  the line based gated convolution text recognition system \cite{flor2020}. The attention was first introduced in Neural Machine Translation (NMT) task \cite{Bahdanau2014,Luong2015}. In a typical NMT task attention is used to provide the importance of each word at given time-step while translating. Other than NMT, attention is also used in speech recognition task \cite{das2019}. Some recent studies have also shown attention to be a promising method used with HTR task. In this study, we have blended the attention mechanism with CTC based NN system to learn and propagate the image features and utilize the benefits obtained from both the techniques The Key contributions of the present study is as follows,
\begin{itemize}
	\item Text recognition system is explained in a end-to-end manner and simplified way.
	\item Popular attention mechanism is merged with existing state-of-the-art HTR techniques.
	\item Two separate NN systems have been taken and merged with attention module to make our observation generalized.
	\item Attention module is merged with Flor et al. and Shi et al. architecture to learn and propagate the image feature efficiently while training of the NN system.
	\item Additionally, Word Beam Search (WBS) decoder \cite{Scheidl2018} has been added  as a post processing step to improve the accuracy in Flor et al system. 
	\item Greedy decoding method is used  with Shi et al. architecture to show the percentage of improvement we observed by adding attention module to existing NN system. 
\end{itemize}
The rest of the paper is organized as follows, section-2 covers the key contribution in the domain of HTR. In section-3, system design is discussed. Section 4 briefed about the experimental setup. Section 5, present the results of this study and comparison with other works. Section 6, covers discussion. At last, conclusions are presented in section 7.

\section{Related Work}
In this section, we have discussed the key contributions in the domain of HTR. We specially focus on the method and techniques involving line level HTR. Early works in this domain has used dynamic programming based methods to segment and identify the words at character level using optimum path finding algorithm \cite{chen99, bellman2015}. Further improvement was observed by using HMM based techniques in HTR \cite{2002vin}. Standard HMM based methods lacked in handling the long sequences of character due to these limitations as per Markov assumption. To improve the accuracy further these models are combined with other basic NN systems. In one such method, a HMM/ANN based architecture is presented in which trigram Language Model (LM) is used for recognition purpose \cite{dreuw2011LM}. Later more advance NN layers systems are studied in connection with the HMMs. In similar study HMM based HTR architecture is used that improves the accuracy of recognition on IAM and RIMES dataset by applying preprocessing steps, discriminative HMM training and discriminative feature extraction. In this, an LSTM network is used for feature extraction,  word and character level LMs are used to improve accuracy further \cite{michal2013LM}. Further, in one such study  the activation function of the  gated units of LSTM is modified to make overall system robust. A combination of HMM and LSTM used as recognizer \cite{patrick2014}. \par Current state-of-the-arts are NN based systems. In one such system, the Convolutional Recurrent Neural Network (CRNN) architecture is introduced that is a combination of CNN and RNN and able to produce state of the results in recognizing sequential objects in scene-text images \cite{shi2015}. Similar to CRNN, one variant of RNNs that is Multidimensional Long Short-Term Memory (MDLSTM) network has been widely researched by community \cite{Graves2008}. MDLSTM architectures provide the recurrence in both of direction (X and Y) along the given image \cite{lalita2022}. Thus, two dimensional data of unconstrained can be processed. Utilizing this nature of MDLSTM many page level recognition systems  of HTR are also proposed. In one such study, the proposed architecture is able to recognize paragraph level text. External segmentation of the paragraph into lines is error prone and this error propagates from  segmentation to recognition. By doing an implicit recognition this study, resolves the issue of error  generating due to poor segmentation. In this study, MDLSTM-RNN is used with attention mechanism. In this work, trigram LM is used that was trained on LOB, Brown and Wellington
corpora \cite{bluche2016}. In similar work,  authors used an attention based model for end to end HTR. In this approach a MDLSTM , CTC and attention based model is gradually trained on from consecutive images of word  to a complete text line. As the size of sentence increases accuracy increases gradually. To  handle the paragraph data, augmentation techniques are used to generate enough training data and model is modified with curriculum learning to adapt to recognition at paragraph level \cite{bluche2016LM}. These networks are complex in nature and require the use of large number of computational resources. But the NN architecture based on convolutional and 1D-LSTM layers are able to learn similar features with very less computational cost \cite{puigcerver2017}. Some notable state-of-the-art systems are only made up of CNN layers or attention techniques without any recurrent layer. \cite{Poznanski2016, Such2018,Coquenet2019,Ptucha2019,Yousef2020,Yousef2020-2,Poulos2021}. 
\section{System Design}
\begin{figure}[!htbp]
	\includegraphics [width=\linewidth, height=19cm]{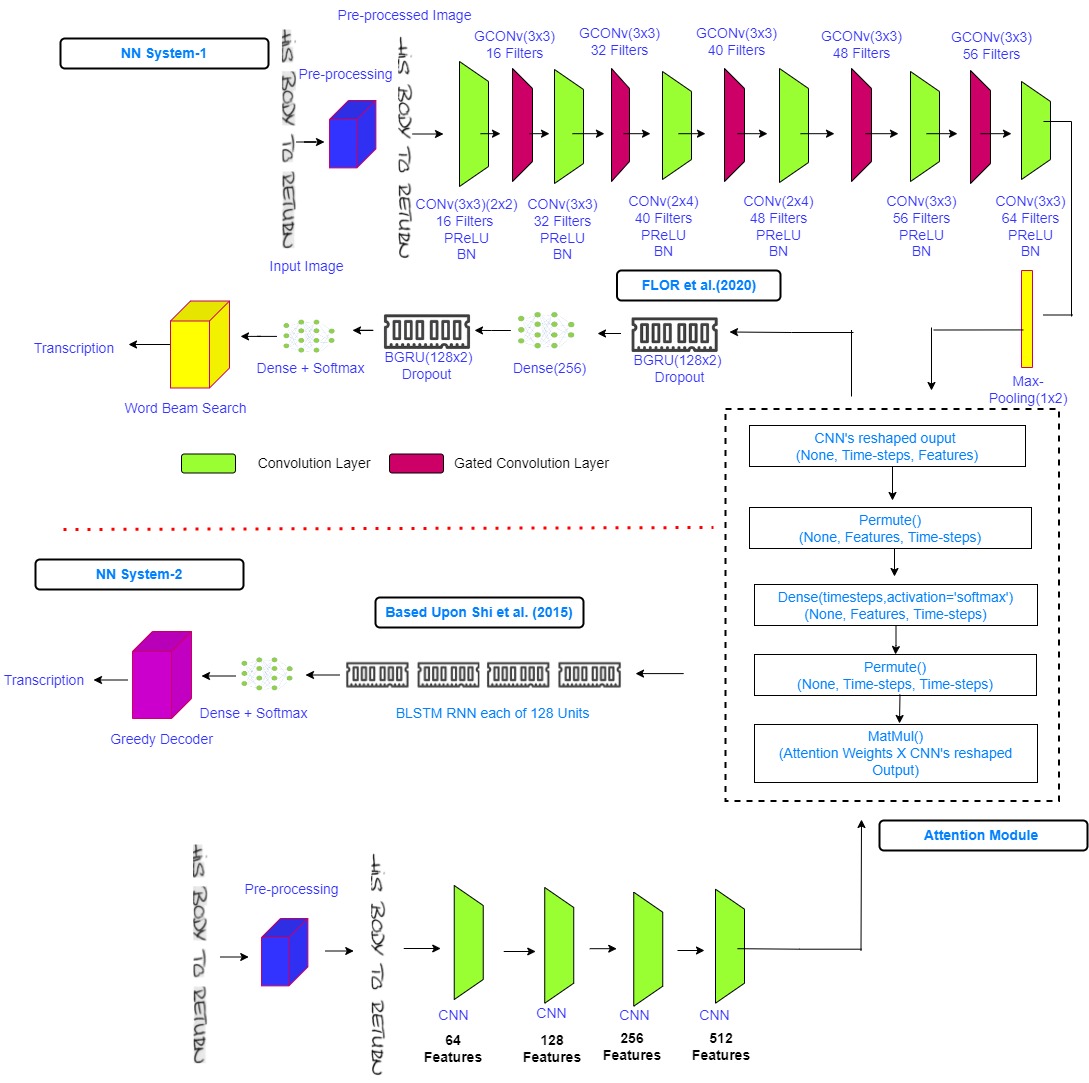}
	\caption {System Architecture}	
	\label{figure:fig-1}
\end{figure}  

In this section, we presented an overview of each  of the module proposed in text recognition system in detail. In this study, we have used modified bahdanau attention \cite{Bahdanau2014} in two state-of-the-art NN systems \cite{flor2020} and \cite{shi2015}. Both of these systems take the input image, extract it features, the propagated features are processed by the attention and RNN layer to produce the occurrence probability of each character at each time step. Figure \ref{figure:fig-1} presents the system architecture in detail. Later in the discussion section, presents more on the position of attention layer in NN model. 
\subsection{Feature Extraction}
In both the systems, a grey scale image is taken as an input and its feature map is produced as an output. A set of convolution layers is used to extract these input image features. The heart of a convolution layer is a convolution operation. A kernel is slide of given input image to produce its feature map. The convolution operation is followed by  Rectified Linear Unit (ReLU) activation function which produces the non-linearity in the NN system. Quick convergence was observed in ReLU comparison to the other activation function of the same class. As shown in Fig. \ref{figure:fig-1} the input image is first pre-processed then this processed image is passed through a set of convolution and fully gated convolution layers in Flor et al. \cite{flor2020} to extract the most relevant features. While in Shi et al. \cite{shi2015} a series of convolution layers with varying kernel sizes are used to extract features of images. 

\subsection{Attention Module}

The CNN output is a four dimension vector which is reshaped to three dimensions (Batch size, time-steps, Features at each time steps). These feature maps ($f_1,f_2...,f_n$) is fed to attention  module as input. The motivation of applying attention is towards getting a more powerful representation using a weighted context vector ($Con_{vec}$). It is computed using Eq. \ref{eq:2},
\begin{equation}\label{eq:2}
	Con_{\mathrm{vec}}^t=\sum_{i=1}^T s_{\mathrm{ti}} f_{\mathrm{i}}^{cnn}
\end{equation}
Where $f_{\mathrm{i}}^{cnn}$ is feature information at $i$\textsubscript{th} time step, $T$ is total number of timesteps and $s_{\mathrm{t}i}$ attention weights at each time steps which are calculated using Eq. \ref{eq:3},
\begin{equation}\label{eq:3}
	s_{\mathrm{i}}=\frac{\exp \left(a_{\mathrm{i}}\right)}{\sum_{k=1}^T \exp \left(a_{\mathrm{k}}\right)}
\end{equation} where $a_{\mathrm{i}}$ is alignment score of $f_i$ feature at $t$ timestep. It is learned using Feed Forward Neural Network (FNN) while training of the NN model. The inputs of different times are aggregated using attention based weighting. Thus, attention mechanism is applied in the time step dimension such that at each step more relevant features will be send to further RNN layers. Before applying the attention mechanism, we have to first permute the reshaped output of CNN. The current output without permutation  will be of form (Batch size, time-steps, Features per time-step) by applying dense layer without permuting indicates that there was no feature exchange due to attention mechanism but that is a false assumption. So, after permute operation, FFN layer is used to get attention probabilities. This operation will be performed for each time step. Thus we obtained a matrix that will contain the weighted attention probabilities at each time step, which further multiplied to feature vectors to obtain the context vector. This context vector will be given to the next layer for further processing.    

\subsection{Recurrent Layers}
 Due to its property of having internal memory recurrent layers are widely used in sequence learning tasks. HTR is a sequence learning problem when we identify the probability of character occurrence given as an input image at each time step. As shown in Fig. \ref{figure:fig-1} the output of attention layer is given to Bidirectional Gated Recurrent Unit (BGRU) in Flor et al. and similarly attention output is given to stacked BLSTM layers in Shi et al. . Both the system used bidirectional stacked layers for processing input in forward and backward direction. 

\subsection{WBS Decoder}
A NN system when trained with CTC loss function, the recurrent layer of such networks produce the character probabilities at each time-step. These probabilities are mapped to final character sequences using various decoding algorithms such as greedy, token passing and WBS decoder. This  was proposed in \cite{Scheidl2018} work. The prefix tree made from available corpus  is internally used by this technique to decide which path to take while decoding at each time step. A Beam is simply one possible character sequence. The Number of beams that takes part in next timestep is equal to beam width except at $t$=0. At final timestep the beam that has the highest probability is selected as final sequence and given as output of the recognizer. We have used Beam Width as 50, processing Mode as 'NGram' while applying this decoder in present study\cite{Scheidl2018}.
\section{Experimenatl Setup}
In this section, we have discussed about the experimental work done in this study. This includes discussion on dataset, preprocessing, data augmentation, evaluation metric and training details. We have used basic model of Flor et al. \cite{flor2020}\footnote[1]{https://github.com/arthurflor23/handwritten-text-recognition}, WBS from \cite{Scheidl2018}\footnote[2]{https://github.com/githubharald/CTCWordBeamSearch}. The NN systems are implemented using Keras package of python.
\subsection{Dataset}
In this study, we evaluate our model on  IAM \cite{IAMMarti2002}  and GW\cite{gw} benchmarked datasets. These datasets contain the pages of handwritten texts. These dataset contains the text images and their transcription at word, line and paragraph level. Each of the both dataset is discussed as below,  
\subsubsection{IAM} 
The IAM dataset contains  the handwritten English text forms used by text recognizers for training and testing purpose. Data present in word, line and paragraph level. In this work we have used the data present at line level with standard split as defined  in Table \ref{table:tab2}. The IAM dataset contains 657 writers, 1539 pages of scanned text, 13353 text lines and 115320 words. All the data is in labelled format. LOB corpus is used to build the IAM dataset. 
\subsection{George Washington-(GW)}
This dataset contains the English letters written by George Washington to their associates in 1755. It has total 20 pages whose data annotated at word level making 5000 words in total. We have used the train, test and validation split as specified in Table \ref{table:tab2}.
\begin{table}[!htbp]

	\caption{Train, validation and test splits of IAM and GW datsets}
	\begin{tabular}{ccccc}
		\textbf{SNo.} &\textbf{Dataset}                                     & \textbf{Train set} & \textbf{Validation set} & \textbf{Test set} \\ \hline
		1 &	\begin{tabular}[c]{@{}c@{}}IAM Dataset\\ (No. of characters=79)\end{tabular}& 6,161                                    & 900                                    & 1,861                                \\ \hline
		2& \begin{tabular}[c]{@{}c@{}}GW Dataset\\ (No. of characters=82)\end{tabular} &325                                     & 168       & 163                                  \\ \hline
	\end{tabular}
	\label{table:tab2}
\end{table}
\subsection{Preprocessing}
\begin{figure}[t]
	\includegraphics [width=\linewidth, height=7cm]{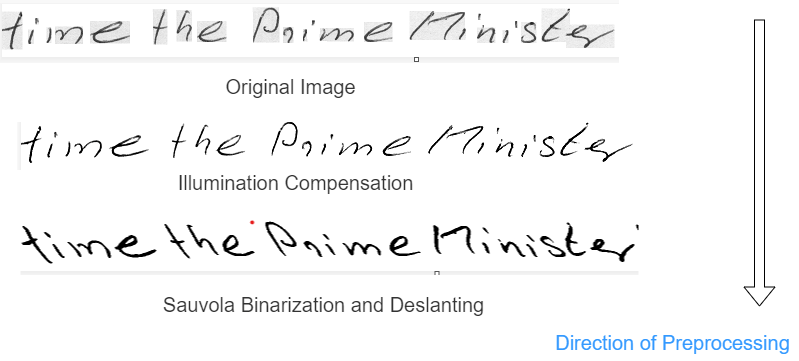}
	\caption {Pre-processing steps}	
	\label{figure:fig-2}
\end{figure}

The preprocessing techniques are used to improve the quality of degraded handwritten text documents. In this study, we have used illumination compensation \cite{CHEN2012}, binarization \cite{SAUVOLA2000} and deslanting \cite{VINCIARELLI2001}  as preprocessing techniques. Figure \ref{figure:fig-2} shows the image of IAM dataset after each preprocess technique. 
\subsection{Evaluation Metric}
The evaluation metric is used to identify how well the proposed system is performing in comparison to earlier studies. We have used the standard evaluation metric Character Error Rate (CER) and Word Error Rate (WER). It is based upon the  Levenshtein distance (LD). It is formulated as follows,
\begin{equation}\label{eq:1}
	WER=\frac{S_{word}+D_{word}+I_{word}}{N_{word}}
\end{equation}  where,
\begin{itemize}
	\item $S_{word}$ is number of substitutions required.
	\item $D_{word}$ is number of deletions required.
	\item and $I_{word}$ is number of insertions required at word level
	\item and $N_{word}$ is total number of characters in ground truth sentence.
\end{itemize}

CER is same as WER the only difference is in CER we work on the character level instead of word level like WER.

\subsection{Data Augmentation}

Data augmentation techniques are used to provide different variations of the samples available for training. For a  NN system to be able to learn properly, right amount of training data is required. The model either can be over fit and under fit based upon the  availability of the data. We have used random morphological and displacement transformations
such as resizing, rotation, image displacement, erosion and dilation.

\subsection{Training Details}
In this section we have discussed the training and testing algorithm used in the present system. Algorithm \ref{algorithm:algo1}
presents the training strategy and explains the details about the attention module. Algorithm \ref{algorithm:algo2} explains the decoding using WBS decoder. In the Shi et all system the training architecture is same expect of number and type of layers but for the decoding purpose we have used best path decoding to produce results in that setup.

\begin{algorithm}[H]

	\SetAlgoLined
\KwIn{line images $I_1,I_2,... I_n$ and ground truth $y_1,y_2,... y_n$ }
	\KwResult{Trained model weights on minimizing the validation loss}
	$epochs$=1000, $batch$=16, $lr$=0.001, $stop\_tolerence$=20, $reduce\_tolerence$=15 \tcp*{initialize the training parameters}
	\SetKwFunction{FMain}{Attention}
	\SetKwProg{Fn}{Function}{:}{}
	\Fn{\FMain{$RNN_{out}$}}{
		{$RNN_{out}$}=Permute(2,1){$RNN_{out}$}\tcp*{permute the time and feature axis}
		{$\alpha_{p}$}=Dense(timestep,softmax)\; \tcp*{Calculate the attention weights}
		{$\alpha_{p}$}=Permute(2,1)	{$\alpha_{p}$}\;
	    {$Context_{vec}$}=Multiply({$RNN_{out}$},{$\alpha_{p}$}) \tcp*{Calculate the context vector}

		\KwRet {$Context_{vec}$} \;
	}
 		\SetKwFunction{FMain}{MAIN}
 	\SetKwProg{Fn}{Function}{:}{}
 	\Fn{\FMain{}}{
        init model() \tcp*{initialize the model framework}
        \For{$i$=1 \KwTo $batch$}{
        	augmentImage($I_i$)\tcp*{Augment text line images}
        	${CNN_i}$=($I_i$) \tcp*{Extracting features of the Image}
        	${Reshape_i}$=Reshape($CNN_i$) \tcp*{Reshaping the output of CNN for further processing}
        	${RNN_{inp-i}}$=Attention(${Reshape_i}$) \tcp*{Attention module}
        	$\hat{y_i}$=RNN(${RNN_{inp-i}}$) \tcp*{Processing RNN Layers}
        	$\hat{y_i}$=Dense(${timestep}$,${Num_{char}}+1$(for CTC blank)) \tcp*{find the character occurrence at each time step}
        	$\delta$\textsubscript{ctc}+=L\textsubscript{ctc}($y_i$,$\hat{y_i}$)\tcp*{compute CTC loss}
        }
        Backward($\delta$\textsubscript{ctc}) \tcp{Updated model weights using back propagation}
        	
 }
		\caption{Training Process}
	\label{algorithm:algo1}
\end{algorithm}

\section*{Explanation}We will discuss training process as in algorithm \ref{algorithm:algo1} in line by line manner as follows\\
 \begin{tabular}{cp{11cm}}
	\textbf{Line 1:}-&  Define training model parameters such as batch size learning rate, early stopping criteria and  total number of epochs.\\	
	\textbf{Line 3:}-& Permute the dimension of the output to enable feature exchange.\\
	\textbf{Line 4:}-& Apply the dense() layer along the timestep.\\
	\textbf{Line 5:}-& Permute back the attention vector\\
	\textbf{Line 6:}-& Obtaining the context vector by multiplying feature with the attention weights for each step. \\
	\textbf{Line 7:}-& Returns the context vector to the main function to be further processed by RNN layers. \\
	\textbf{Line 9:}-& Load the NN model.\\
	\textbf{Line 11:}-& For a given batch augment the preprocessed image.\\
	\textbf{Line 12:}-& Extract the features of the image using series of convolutions and gated convolutions operations.\\
	\textbf{Line 13:}-& Converting 4D feature maps to 3D vectors to be further processed by attention and RNN layers.\\
	\textbf{Line 14:}-&  Applied the attention as defined above and obtained the context vector.\\
	\textbf{Line 15:}-&  Find the predicted character occurrence from the output of RNN \\
	\textbf{Line 16:}-&  Map the output of RNN to the number of characters of dataset + 1  for the sequence prediction. \\
	\textbf{Line 17:}-&  Compute the CTC loss from predicted and actual character sequence. \\
	\textbf{Line 19:}-&  Based upon the Loss value train the NN system using backpropogation.	   
\end{tabular}
\begin{algorithm}[]
	\SetAlgoLined
	\KwIn{ Text line image $I$, $E_{test_{corpus}}$, $E_{chars}$, $E_{wordchars}$}
	\KwResult{Prediction of image text with CER and WER}
	$BW$=50, $mode$='NGrams', $smooth$=0.01 \tcp*{initialize the WBS decoding parameters}
	initNNModel() \tcp*{Loading of the trained model with all the layers}
	output=Modelpredict($I$) \tcp*{Predict the text in the Image}
	$\hat{y}$= Decoder($BW$,$mode$, $smooth$=0.01,$D_{test_{corpus}}$, $D_{chars}$, $D_{wordchars}$) \tcp*{Applying WBS decoding algorithm}
	$CER$,$WER$= accuracy($y$,$\hat{y}$) \tcp*{compute CER and WER}
	\caption{Prediction process}
	\label{algorithm:algo2}
\end{algorithm}
\section*{Explanation} We will discuss prediction process as in algorithm \ref{algorithm:algo2} in line by line manner as follows,

\begin{tabular}{lp{10cm}}
	
	\textbf{Line 1:}-&  Input text line image $I$. First, initialize the parameters of WBS decoding\\
	\textbf{Line 2:}-& Build the model.\\
	\textbf{Line 3:}-&  Process the image as per the trained model.\\
	\textbf{Line 4:}-& RNN's output dimensions are swapped as per predefined input accepted by WBS decoder.\\
	\textbf{Line 5:}-& Computation of character occurrence using WBS decoding algorithm.\\
	\textbf{Line 6:}-& Estimate the  accuracy of the model on test images.
\end{tabular}    	
\section{Results and Comparison}
In this section, the results obtained in the present study have been discussed and compared with the other state-of-the art methods. This HTR system recognizes the handwritten text on line level so the results are compared with other state-of-the-art line level systems. We are able to achieve 4.12 \%CER and 9.72\% WER on IAM dataset  and 7.07 \% CER and 16.14\% WER on GW dataset having Flor et. al. as our based model and WBS as a decoding algorithm. We have also implemented Shi et. al. architecture and merged the attention module with that. Similar improvements were observed in that architecture reported in Table \ref{table:tab5}. The greedy decoder is used in this system. The given attention module is providing  23.27\% improvement from basic model.

 \begin{table}[h]
 	\vspace{-4em}
 	\ra{1}
 	\caption{Comparison of present work with other state-of-the-art line level works on IAM Dataset}
 	\centering 
 	\begin{tabular}{lllll}
 		\textbf{SNo.}	& \textbf{Reference}  & \begin{tabular}[c]{@{}c@{}}\textbf{Method}\\ \textbf{Technique}\end{tabular}   & \textbf{CER}         & \textbf{WER}         \\ \hline
 		1 & Puigcerver et al. \cite{puigcerver2017}	& \begin{tabular}[c]{@{}c@{}}CNN + LSTM +\\ CTC\end{tabular} & 4.4                    & 12.2                     \\ \hline
 		2 & Chowdhury et al.\cite{Chowdhury2018} & \begin{tabular}[c]{@{}c@{}}CNN + BLSTM +\\ LSTM\end{tabular}       & 8.1                   & 16.7                     \\ \hline
 		3 & Michael et al. \cite{Michael2019} & \begin{tabular}[c]{@{}c@{}}CNN + LSTM +\\ Attention\end{tabular}  & 4.87                  & -                  \\ \hline
 		4 &  Kang et al. \cite{Kang2020} & Transformer      & 4.67                 & 15.45                \\ \hline
 		5 & Yousef et al.\cite{Yousef2020} & CNN + CTC       & 4.9     & -               \\ \hline
 		6 & Flor et al. \cite{flor2020} &\begin{tabular}[c]{@{}c@{}}CNN + BGRU +\\ CTC\end{tabular} & 3.72 & 11.18 \\ \hline
 		7 &	 Present Work & \begin{tabular}[c]{@{}c@{}}CNN + Attention +\\ BGRU + CTC\end{tabular}       & \textbf{4.15}              & \textbf{9.72}              \\ \hline

 	\end{tabular}
 	\label{table:tab3}
 \end{table}
\begin{table}[]
	\ra{1.5}
	\caption{Comparison of present work with other state-of-the-art line level works on GW Dataset}
	\centering 
	\begin{tabular}{lllll}
		\textbf{SNo.} &	\textbf{Reference} & \textbf{Method/Technique}  & \textbf{CER}         & \textbf{WER}         \\ \hline
		1 & Toledo et al. \cite{Toledo2017} & CNN + BLSTM + CTC   &  7.32                   & -                   \\ \hline
		2 & Almazan et al. \cite{Almazan2014}  &   Word Embedding  &  17.40                  &    -          \\ \hline
		3 & Fischer et al. \cite{Fischer2012} &   HMM + RNN   &  20  & -   \\ \hline
		4 & Present Work &  \begin{tabular}[c]{@{}c@{}}CNN + Attention +\\ BGRU + CTC\end{tabular}  &  \textbf{7.07}         & \textbf{16.14}                   \\ \hline
	\end{tabular}
	\label{table:tab4}
\end{table}
\begin{table}[!htbp]
	\ra{1.5}
		\caption{Similar NN System to Shi et. al. recreated and attention module applied CNN as in Flor et. al. except using greedy decoder instead of WBS}
	\centering 
	\begin{tabular}{ll}
		\textbf{Architecture}   & \textbf{CER(in \%)} \\ \hline
		\multicolumn{1}{l}{Model Based Upon Shi et. al} & 11.00               \\ \hline
		Model Based Upon Shi et al + Attention          & 8.44                \\ \hline
	\end{tabular}
	\label{table:tab5}
\end{table}
\section{Discussion}
\begin{figure}[!htbp]
	\includegraphics [width=\linewidth, height=6cm]{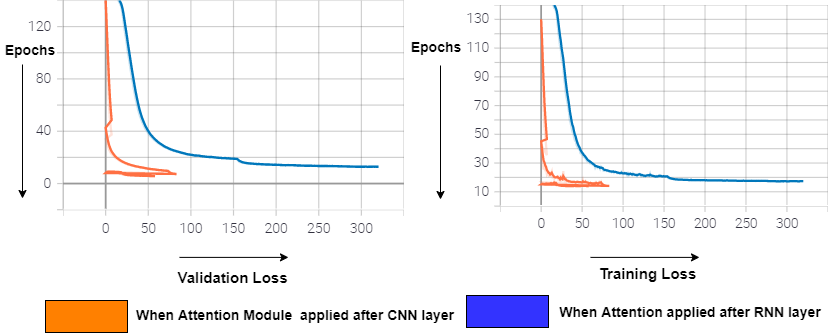}
	\caption {Comparison of Training and Validation loss at different places of attention module}	
	\label{figure:fig-3}
\end{figure}

In the proposed study, attention mechanism is helped in identifying relevant features in given input image. We have experimentally found out two possible positions where this attention block can be plugged in. Through extensive experiments and as per the graphs shown in Fig. \ref{figure:fig-3} it is evident that the use of attention mechanism before the recurrent layers and after the CNN layer  helps the model to converge quicker with a better accuracy.  As shown in the Fig. \ref{figure:fig-3} with the same hyper-parameters for training the model  learns better and converge quickly when attention module applied after CNN layers. 
\section{Conclusion}
In the present study, we have merged attention with two state-of-the-art NN systems that is Flor et al. and small version of Shi et al.
we are able to achieve  4.15\% CER and 9.72\% WER on IAM dataset, 7.07\%  CER and 16.14\% WER on GW dataset. We have also observed the 23.17\% improvement in CER from the base model  by applying attention module in NN system similar to Shi et. al. system. The accuracies obtained after applying attention module favours our hypothesis that  attention helps in learning the image features better. The position of applying the attention module is a critical step to consider which we addressed in the discussion section. In future, we will work on extended this work for page or paragraph level. 

\textbf{Acknowledgements} This research is funded by Government of India, University Grant Commission, under Senior Research Fellowship scheme.

%% The Appendices part is started with the command \appendix;
%% appendix sections are then done as normal sections
%% \appendix

%% \section{}
%% \label{}

  \bibliographystyle{elsarticle-num} 
  \bibliography{bibliography}

\end{document}